\pdfoutput=1
\documentclass[11pt]{article}
\usepackage[T1]{fontenc}
\usepackage[preprint]{acl}
\usepackage{times}
\usepackage{latexsym}
\usepackage{tabularx}
\usepackage{multirow}
\usepackage{booktabs}
\usepackage{setspace} 
\usepackage{tcolorbox}
\usepackage{lipsum} 
\usepackage{graphicx}
\usepackage{amssymb}
\usepackage{float}
\usepackage{arydshln}
\usepackage{hyperref}
\usepackage{microtype}
\usepackage{inconsolata}
\usepackage{enumitem}

\title{How Well Do Large Reasoning Models Translate?\\A Comprehensive Evaluation for Multi-Domain Machine Translation}

\author{
Yongshi Ye\textsuperscript{1,3,}\thanks{\,\,Equal contribution.},
Biao Fu\textsuperscript{2,3,}\footnotemark[1]\textsuperscript{,}\thanks{\,\,Corresponding author.},
Chongxuan Huang\textsuperscript{2,3},
Yidong Chen\textsuperscript{2,3},
Xiaodong Shi\textsuperscript{2,3,}\footnotemark[2]
\\[0.5em]
\textsuperscript{1}Institute of Artificial Intelligence, Xiamen University \\
\textsuperscript{2}School of Informatics, Xiamen University \\
\textsuperscript{3}Key Laboratory of Digital Protection and Intelligent Processing of Intangible Cultural \\ Heritage of Fujian and Taiwan (Xiamen University),Ministry of Culture and Tourism\\
\texttt{\{yeyongshi,biaofu\}@stu.xmu.edu.cn,mandel@xmu.edu.cn}
}

\begin{document}
\maketitle
\begin{abstract} 
Large language models (LLMs) have demonstrated strong performance in general-purpose machine translation, but their effectiveness in complex, domain-sensitive translation tasks remains underexplored. Recent advancements in Large Reasoning Models (LRMs), raise the question of whether structured reasoning can enhance translation quality across diverse domains. In this work, we compare the performance of LRMs with traditional LLMs across 15 representative domains and four translation directions. Our evaluation considers various factors, including task difficulty, input length, and terminology density. We use a combination of automatic metrics and an enhanced MQM-based evaluation hierarchy to assess translation quality. Our findings show that LRMs consistently outperform traditional LLMs in semantically complex domains, especially in long-text and high-difficulty translation scenarios. Moreover, domain-adaptive prompting strategies further improve performance by better leveraging the reasoning capabilities of LRMs. These results highlight the potential of structured reasoning in MDMT tasks and provide valuable insights for optimizing translation systems in domain-sensitive contexts\footnote{The evaluation results are available at \url{https://github.com/WingseeYe/TransEval-LRM}.}. 
\end{abstract}

\section{Introduction}

Large language models (LLMs) have been widely applied to machine translation (MT) and demonstrate superior performance in general-domain settings~\cite{mao-etal-2024-gpteval, peng-etal-2023-towards, jiao2023chatgpt, hendy2023good, siu2023chatgpt}. 
However, LLMs face persistent challenges in multi-domain MT (MDMT), including instability in domain disambiguation, poor alignment with domain-specific terminology and style, and limited cross-domain generalization~\cite{zheng2024fine,hu-etal-2024-large-language, man2025dmdteval}. 

Recently, a new class of reasoning-oriented LLMs known as Large Reasoning Models (LRMs) has emerged, including models such as OpenAI-o1~\cite{jaech2024openai} and DeepSeek-R1~\cite{deepseekai2025deepseekr1incentivizingreasoningcapability}. These models incorporate structured generation mechanisms such as multi-step planning, intermediate verification, and self-refinement, and have achieved strong results in complex tasks such as code generation~\cite{zhang2024o1codero1replicationcoding, quan2025codeelo, zhang2025codecriticbench} and mathematical reasoning~\cite{zhao2025promptcot, huang2025math}. 
However, it remains unclear whether these reasoning models can generalize to domain-sensitive MT—particularly MDMT—and effectively address its unique challenges, including structural complexity, long-range dependencies, terminology alignment, domain variation, and stylistic consistency.
Unfortunately, this question has received little attention, and no prior work has systematically examined the translation capabilities of LRMs in multi-domain settings. 

In this work, we aim to fill this gap by systematically evaluating the translation capabilities of LRMs across diverse domains.
To this end, we conduct a comprehensive comparison between traditional LLMs and LRMs across 15 representative domains and four translation directions.
Specifically, our evaluation spans 11 domains in the German$\Leftrightarrow$English setting and 12 domains in the Chinese$\Leftrightarrow$English setting, covering a wide range of content—from general-purpose domains such as conversation and news to highly specialized areas like law, medicine, and literature. 
Our evaluation combines automatic metrics (BLEU, COMET, CometKiwi) with an LLM-powered scoring protocol based on the Multidimensional Quality Metrics (MQM) hierarchy, allowing us to analyze how reasoning capacity and domain control interact across diverse translation conditions.

Based on our comprehensive evaluation, we summarize the following key findings:
\textbf{First}, LRMs achieve higher semantic metrics (COMET, CometKiwi) in open-domain settings (e.g., Conversation, News), but lower BLEU in terminology-intensive domains. 
\textbf{Second}, MQM evaluations on De$\Rightarrow$En show that LRMs perform well in semantic accuracy and fluency, but are weaker in stylistic consistency. 
In terminology accuracy, LRMs outperform traditional LLMs in Zh$\Rightarrow$En but underperform in De$\Rightarrow$En, suggesting that model suitability may vary across language pairs.
\textbf{Third}, LRMs demonstrate superior adaptability in document-level translation. Their structured reasoning mechanisms help model longer-range dependencies and cross-sentence coherence, yielding more consistent outputs for long-form texts.
\textbf{Fourth}, we find that incorporating domain-aware prompts can significantly improve their performance. Prompting strategies that encourage implicit domain inference (rather than static domain label) lead to better semantic and stylistic alignment.
\textbf{Fifth}, we show that LRMs outperform traditional LLMs on complex translation tasks. Their structured reasoning capabilities allow them to maintain semantic integrity even as translation complexity increases.

Overall, our study not only benchmarks the translation capabilities of LRMs across diverse domains, but also sheds light on their unique strengths and limitations compared to traditional LLMs, offering valuable insights for future research. 

\section{Related Work}

\subsection{Large Reasoning Models}

Enhancing the reasoning capabilities of LLMs has become a critical step toward general intelligence. OpenAI’s o1~\cite{jaech2024openai} has demonstrated strong performance on complex reasoning tasks by aligning structured thinking via reinforcement learning (RL). 
Building on this foundation, the “Journey Learning” paradigm~\cite{qin2024o1replicationjourneystrategic} encourages models to emulate human-like exploration strategies, including trial-and-error and reflection, while subsequent work distills o1-style chains-of-thought (CoT) through supervised fine-tuning (SFT) to improve reasoning stability and depth~\cite{huang2024o1replicationjourney, wang2025ma, ma2025sorftissueresolvingsubtaskoriented, yu2025longshortchainofthoughtmixturesupervised, trung-etal-2024-reft}. 
RL-based strategies have since been widely adopted. DeepSeek-R1~\cite{deepseekai2025deepseekr1incentivizingreasoningcapability} applies multi-stage RL training and cold-start Long CoT data to enhance reasoning.
Gemini2.0-Flash-Thinking~\cite{googledeepmind2024gemini20flashthinking} extends context windows to support long-range reasoning. More broadly, models like Marco-o1~\cite{zhao2024marcoo1openreasoningmodels} aim to align reasoning capabilities for general-purpose, open-ended problem-solving.

In the context of MT, reasoning-oriented models remain in their early stages but have demonstrated promising potential~\cite{liu2025newtrendsmodernmachine}. 
Recent approaches such as DRT~\cite{wang2025drtdeepreasoningtranslation}, R1-T1~\cite{he2025r1}, DeepTrans~\cite{wang2025deepreasoningtranslationreinforcement}, and MT-R1-Zero~\cite{feng2025mt} employ Long CoT-based SFT and/or RL to explicitly activate the reasoning capabilities of LLMs for translation tasks. However, these methods have mainly focused on general-domain or single-domain settings, and a systematic evaluation of reasoning capabilities in MDMT remains lacking.

\subsection{Evaluation for LLM-based MDMT}

LLMs have shown strong performance in general-domain MT, prompting growing interest in their applicability to domain-specific settings~\cite{qian-etal-2024-large, huang-etal-2024-lost, wang-etal-2023-document-level, zhu-etal-2024-multilingual}. Recent work has extended evaluation to diverse genres and specialized domains. In literary translation, studies based on paragraph-level and full-novel corpora~\cite{thai-etal-2022-exploring, fonteyne-etal-2020-literary, zhang2024good} report fluent outputs but persistent issues in stylistic fidelity, discourse coherence, and pragmatic nuance. In technical domains such as science, law, and IT, evaluations highlight challenges with terminological accuracy and structural rigor~\cite{eschbach2024exploring, kleidermacher2025sciencelanguagesassessingllm}. 
Benchmarking efforts~\cite{yan2024benchmarkinggpt4humantranslators, yan2024gpt4vshumantranslators} further reveal limited generalization in low-resource and specialized domains. In structured contexts like e-commerce, targeted fine-tuning improves fluency and semantic similarity~\cite{gao2024llms}.

Although these studies have provided detailed evaluations of traditional LLMs in domain-specific translation, recent advances in LRMs raise the question of whether such models are better suited to the challenges of domain-sensitive MT.
\citet{chen2025evaluatingo1likellmsunlocking} have recently evaluated LRMs on general MT tasks; however, their effectiveness in the more challenging MDMT remains underexplored.
In this work, we fill this gap with a comprehensive evaluation across diverse domains.

\section{Evaluation Setting}
\noindent \textbf{Dataset.}
To systematically evaluate the performance differences between traditional LLMs and LRMs in MDMT, we collect test sets for both German$\Leftrightarrow$English and Chinese$\Leftrightarrow$English directions, covering representative domains for each.

For German$\Leftrightarrow$English, the evaluation spans 11 domains. Five domains—Medical, Law, IT, Koran, and Subtitles—are sourced from the publicly available Multi-Domain dataset~\cite{aharoni-goldberg-2020-unsupervised}. The Mixed domain is drawn from the WMT22 General Machine Translation Task~\cite{kocmi-etal-2022-findings}. Its subdomains include News, Social, E-commerce, and Conversation. Biomedical data is derived from the WMT18 and WMT19 Biomedical Machine Translation Task~\cite{neves-etal-2018-findings,bawden-etal-2019-findings}.

For Chinese$\Leftrightarrow$English, the evaluation covers 12 domains. The CommonSense and Culture domains are taken from the CommonMT~\cite{he-etal-2020-box} and CAMT~\cite{yao-etal-2024-benchmarking} datasets, respectively, while the Literary domain comes from the WMT23 Literary Machine Translation Task~\cite{wang-etal-2023-findings}. The general-purpose Mixed domain, along with its subdivisions (News, Social, E-commerce, and Conversation), is again sourced from the WMT22 General Machine Translation Task~\cite{kocmi-etal-2022-findings}. Biomedical data is also collected from WMT18~\cite{neves-etal-2018-findings} and WMT19~\cite{bawden-etal-2019-findings} Biomedical Machine Translation Task. Additionally, the Laws, News, Science, and Subtitles domains are sourced from the UM-Corpus~\cite{tian-etal-2014-um}.

The selected domains span a wide range of linguistic and contextual characteristics, including high-resource domains such as news; terminology-intensive areas like IT, law, medicine, biomedicine, and e-commerce; low-resource domains such as Koranic texts; context-sensitive areas like culture and commonsense; and stylistically demanding content such as literature. This design ensures high diversity in domain coverage, textual style, and semantic complexity, enabling a comprehensive and multidimensional evaluation of model performance. Detailed data statistics are provided in Appendix~\ref{appendix:dataset}.

\begin{figure*}[th] 
    \centering
    \includegraphics[width=\textwidth]{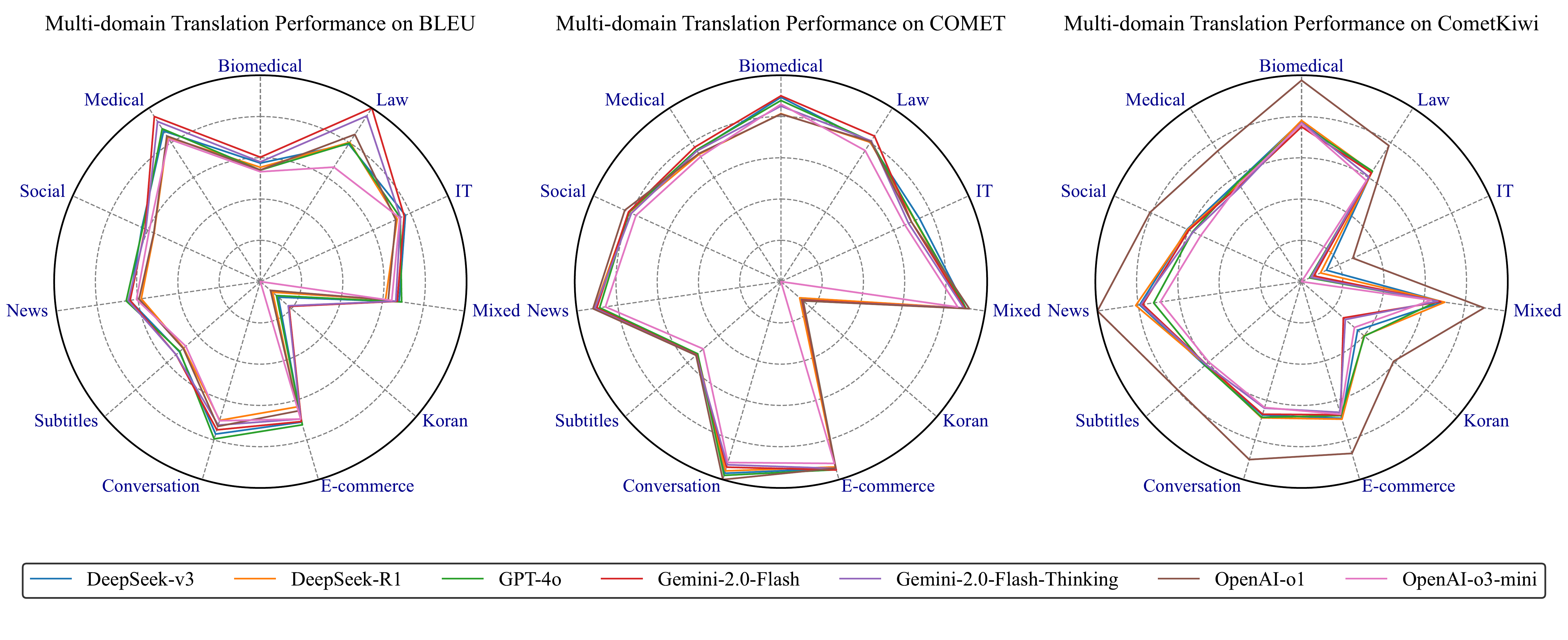}
    \caption{Multi-Domain En$\Leftrightarrow$De Translation Performance Comparison, showing averaged BLEU, COMET, and CometKiwi scores for both directions, with distinct colors representing different LLMs. }
    \label{fig:de-en-radar}
\end{figure*}

\begin{figure*}[th]  
    \centering
    \includegraphics[width=\textwidth]{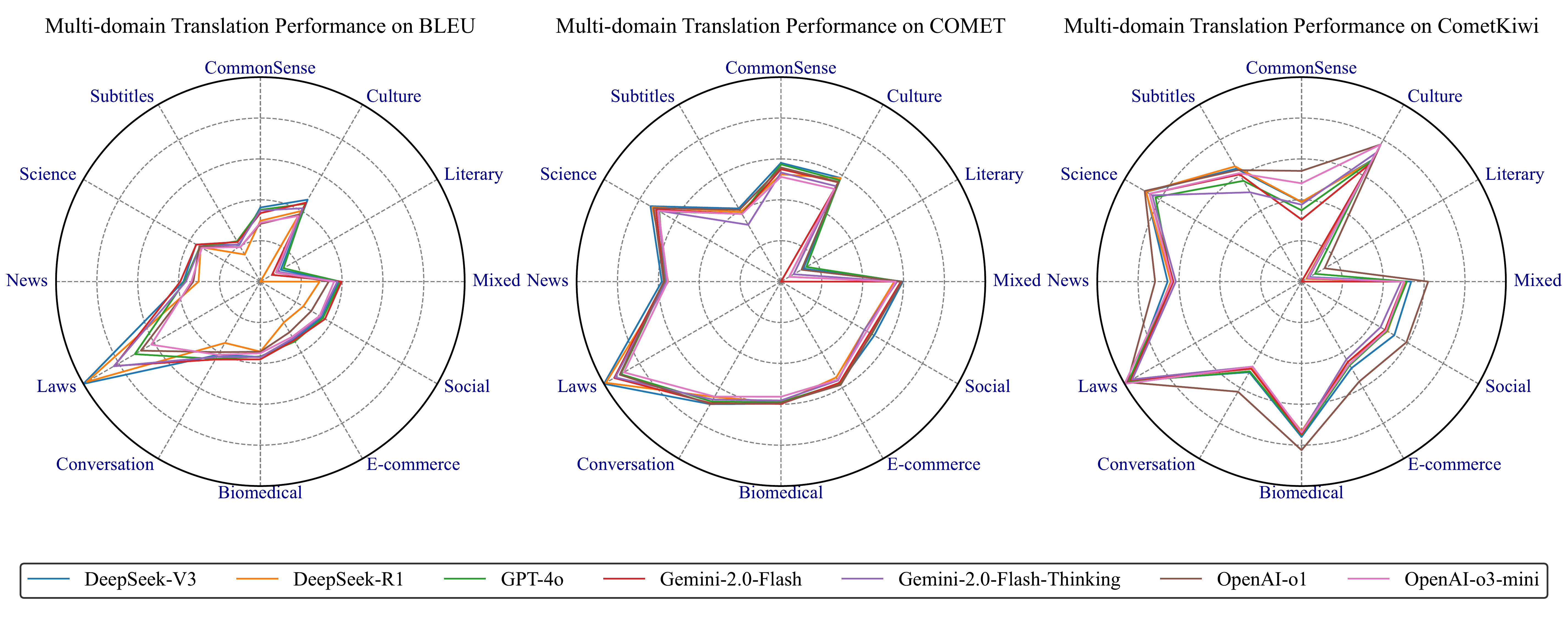}
    \caption{Multi-Domain En$\Leftrightarrow$Zh Translation Performance Comparison, showing averaged BLEU, COMET, and CometKiwi scores for both directions, with distinct colors representing different LLMs. }
\end{figure*}

\noindent \textbf{Metrics.}
In this experiment, we evaluate the performance of reasoning-based and traditional LLMs in MDMT using three automated metrics: BLEU\footnote{\url{https://github.com/mjpost/sacrebleu}}~\cite{post-2018-call}, COMET\footnote{Unbabel/wmt22-comet-da}~\cite{rei-etal-2022-comet}, and CometKiwi\footnote{Unbabel/wmt22-cometkiwi-da}~\cite{rei-etal-2022-CometKiwi}. BLEU emphasizes lexical-level overlap between the hypothesis and reference, while COMET and CometKiwi provide a semantically grounded evaluation that better correlates with human judgment, especially in complex tasks. Specifically, COMET, as a reference-based metric, measures the discrepancy between a model's output and the reference translation, while CometKiwi, as a reference-free metric, provides an objective evaluation when the reference translation quality is low. In addition, we apply the MQM\footnote{\url{https://themqm.org}}~\cite{lommel2014multidimensional,klubivcka2018quantitative}, which we adapt to meet the demands of multi-domain evaluation by incorporating domain-specific features and characteristics of LLMs. A detailed description of the prompt and MQM hierarchy is provided in Appendix~\ref{appendix:mqm}. Designed with input from translation professionals, the hierarchy uses LLMs as the automated evaluator to annotate errors and assign severity levels. This standardized protocol ensures that the evaluation results are interpretable and adaptable to different domains.

\noindent \textbf{Model.}
We group the evaluated models into two categories. 
\textbf{Traditional LLMs} include GPT-4o~\cite{openai2024gpt4ocard}, DeepSeek-V3~\cite{deepseekai2025deepseekv3technicalreport}, and Gemini-2.0-Flash~\cite{googledeepmind2024gemini20}. \textbf{LRMs} consist of OpenAI-o1~\cite{jaech2024openai}, OpenAI-o3-mini~\cite{openai2025o3mini}, DeepSeek-R1~\cite{deepseekai2025deepseekr1incentivizingreasoningcapability}, and Gemini-2.0-Flash-Thinking~\cite{googledeepmind2024gemini20flashthinking}.
All models are accessed via official APIs provided by DeepSeek, OpenAI, and Gemini\footnote{The OpenAI, DeepSeek, and Gemini models used in this study are accessed via the following APIs: gpt-4o-2024-11-20, o1-2024-12-17, and o3-mini-2025-01-31 for OpenAI; deepseek-chat-2024-12-26 and deepseek-reasoner for DeepSeek; and gemini-2.0-flash and gemini-2.0-flash-thinking-exp-2025-01-21 for Gemini.}. 

\section{Experiments}

\subsection{Overall Analysis}
\label{all_exp_results}

\noindent \textbf{En$\Leftrightarrow$De}
In both En$\Rightarrow$De and De$\Rightarrow$En directions, LRMs perform better on the COMET metric in domains such as Conversation, E-commerce, News, and Social, which are characterized by natural language, diverse expression, and low contextual dependency. This suggests that their generation mechanism, aligned with human preferences, is better suited for open-domain and flexible-context translation tasks, enabling more accurate semantic representation. However, in structurally rigid and terminology-intensive domains such as Medical, Law, IT, and Koran, LRMs achieve lower COMET scores than traditional LLMs. 
This may be attributed to their tendency to prioritize semantic plausibility over strict adherence to domain-specific terminology.

Moreover, LRMs may "overthink"~\cite{chen2025overthinking} or over-interpret specialized terminology, leading to semantically coherent but lexically divergent translations.
Notably, their performance on the reference-free CometKiwi metric remains strong, indicating good semantic consistency, especially for OpenAI-o1, which outperforms traditional LLMs on the CometKiwi across all domains.

\noindent \textbf{En$\Leftrightarrow$Zh}
Similar patterns are observed in both En$\Rightarrow$Zh and Zh$\Rightarrow$En directions.
LRMs demonstrate strong semantic alignment in domains like E-commerce and Conversation, where language is flexible and user-focused, yielding high COMET scores. However, they perform less effectively in domains such as Law and Science, which require precise, terminology-heavy translation. 
Additionally, in the Zh$\Rightarrow$En Culture domain (Table~\ref{tab:result_culture_mt}), DeepSeek-V3 performs better, likely due to its Chinese-centric pretraining, which enhances its ability to capture linguistic and cultural nuances more effectively compared to other models. These findings suggest that translation quality in this setting is shaped by both reasoning paradigms and pretraining data alignment. While LRMs consistently top CometKiwi across domains, further gains in terminological accuracy and syntactic precision remain critical.

\noindent \textbf{Overall Perspective}
Across both language pairs, traditional LLMs consistently outperform LRMs on BLEU, particularly in domains such as law, medicine, and IT where strict formatting and terminological consistency are essential. 
This performance gap may be attributed to differences in training paradigms.
Traditional LLMs may have conducted SFT on parallel translation data from these domains, which encourages the model to directly imitate the reference translation, aligning well with how BLEU evaluates translations.
In contrast, LRMs are typically optimized through RL, which prioritizes global semantic consistency and alignment with human preferences, leading to better performance on semantic metrics such as COMET and CometKiwi, but reduced lexical overlap with the reference.
This leads to stronger performance on semantic metrics such as COMET and CometKiwi, but results in lower lexical overlap with references and thus reduced BLEU scores.
Importantly, lower BLEU scores for LRMs in terminology-rich domains do not always reflect actual translation errors. These models often generate semantically equivalent terms or valid paraphrases that diverge from the reference but remain correct. BLEU, being reference-bound and surface-focused, fails to capture such variation. 
This observation highlights the necessity of multi-dimensional evaluation. To obtain a more complete picture of translation quality, we conduct fine-grained evaluation using the MQM hierarchy in the next section.

% MQM 
\subsection{MQM Evaluation}
To systematically evaluate translation errors across domains, we adopt the MQM hierarchy and employ DeepSeek-V3 as an automatic scoring agent to ensure consistent annotation quality. A structured prompt guides the model to identify up to five major errors per sentence, and scores are computed using severity weights proposed by~\citet{freitag-etal-2021-experts}. This approach balances scalability and consistency, allowing fine-grained comparison across models on seven key dimensions: Accuracy, Fluency, Style, Terminology, Others, Source Error, and Non-translation Error. In MQM score calculation and error category statistics, Source Error are excluded~\cite{freitag-etal-2021-experts}.

\begin{table}[t]
\centering
% \small
\resizebox{\linewidth}{!}{%
\begin{tabular}{l c c c c c c}
\toprule
\multirow{2}{*}{Model} & \multicolumn{6}{c}{MQM Score ($\downarrow$)} \\
\cmidrule(lr){2-7}
 & IT & Koran & Medical & Law & Subtitles & Average\\
\midrule
GPT-4o & 3.14 & 4.56 & 3.24 & 3.74 & 3.44 & 3.62 \\
DeepSeek-V3 & \textbf{2.91} & 4.62 & 3.30 & 3.89 & \textbf{3.38} & 3.62 \\
Gemini-2.0-Flash & 3.09 & 4.68 & \textbf{3.15} & \textbf{3.55} & \textbf{3.38} & \textbf{3.57} \\
\hdashline
OpenAI-o1 & 3.11 & 4.74 & 3.32 & 3.86 & 3.49 & 3.70 \\
DeepSeek-R1 & 2.92 & \textbf{4.55} & 3.30 & 3.88 & 3.40 & 3.61 \\
OpenAI-o3-mini & 3.16 & 5.22 & 3.43 & 4.59 & 3.62 & 4.00 \\
Gemini-2.0-Flash-Thinking & 3.15 & 4.84 & 3.28 & 3.69 & 3.41 & 3.67 \\
\bottomrule
\end{tabular}%
}
\caption{MQM scores for different models across various domains in the Multi-Domain dataset (De$\Rightarrow$En).}
\label{tab:mqm_avg_scores}
\end{table}

\begin{table}[t]
\centering
\resizebox{\linewidth}{!}{%
\begin{tabular}{l ccccc}
\toprule
\textbf{Model} & \textbf{Acc. (\%)} & \textbf{Style (\%)} & \textbf{Fluency (\%)} & \textbf{Term. (\%)} & \textbf{Non-trans (\%)} \\
\midrule
GPT-4o & 50.78 & 30.58 & 7.03 & 11.38 & 0.23 \\
DeepSeek-V3 & 50.72 & 30.51 & 6.84 & 11.83 & 0.09 \\
Gemini-2.0-Flash & 52.86 & \textbf{29.43} & 6.87 & \textbf{10.68} & 0.16 \\
\hdashline
OpenAI-o1 & 50.95 & 29.58 & 7.10 & 12.22 & 0.15 \\
DeepSeek-R1 & 50.50 & 30.83 & 6.68 & 11.91 & \textbf{0.08} \\
OpenAI-o3-mini & \textbf{49.97} & 30.48 & 7.19 & 12.24 & 0.12 \\
Gemini-2.0-Flash-Thinking & 52.54 & 29.64 & \textbf{6.38} & 11.29 & 0.15 \\
\bottomrule
\end{tabular}%
}
\caption{Error category percentages for different models, including accuracy (Acc.), style, fluency, terminology (Term.), and non-translation error (Non-trans).}
\label{tab:model_error_category_percentages}
\end{table}

As shown in Table~\ref{tab:mqm_avg_scores}, traditional LLMs achieve lower overall MQM scores, showing higher consistency and precision.
Interestingly, LRMs achieve lower MQM scores in the Koran domain, which may benefit from their explicit CoT reasoning paradigm when handling semantically ambiguous or stylistically atypical text. 
However, their performance lags behind in terminology-intensive domains such as Medical and Law. 
Further analysis in Table~\ref{tab:model_error_category_percentages} reveals that LRMs have slightly lower error rates in Accuracy, Fluency, and Non-translation Error, suggesting strong capabilities in semantic similarity and sentence structure. In contrast, they exhibit higher error rates in Style and Terminology—particularly OpenAI-o3-mini and DeepSeek-R1—reflecting weaker handling of domain-specific consistency and stylistic coherence. 

These findings highlight the need for future work to further strengthen LRMs' capabilities in deep domain understanding, contextualized terminology selection, and adaptation to distinct domain-specific translation styles, to enhance their robustness in domain-diverse translation scenarios.

\begin{table}[t]
\centering
\small
% \resizebox{.5\textwidth}{!}{%
\begin{tabular}{lcc}
\toprule
\multirow{2}{*}{Model} & \multicolumn{2}{c}{Term Acc. (\%)} \\
\cmidrule(lr){2-3}
& Zh$\Rightarrow$En & De$\Rightarrow$En \\
\midrule
GPT-4o & 52.95 & 37.67 \\
DeepSeek-V3 & 51.84 & 37.92 \\
Gemini-2.0-Flash & 52.21 & \textbf{38.55} \\
\hdashline
OpenAI-o1 & 52.92 & 37.42 \\
DeepSeek-R1 & \textbf{53.20} & 37.83 \\
OpenAI-o3-mini & 45.59 & 37.38 \\
Gemini-2.0-Flash-Thinking & 52.67 & 37.62 \\
\bottomrule
\end{tabular}
% }
\caption{Terminology accuracy (\%) for Zh$\Rightarrow$En and De$\Rightarrow$En translation tasks.}
\label{tab:termacc}
\end{table}

\subsection{Terminological Accuracy}
In the previous section, we observed that LRMs underperform traditional LLMs in terminology-heavy domains such as law and medicine. This observation prompted a focused investigation into their terminology translation capabilities. To this end, we conducted an evaluation based on the WMT23 Terminology Shared Task~\cite{semenov-etal-2023-findings}, which provides bilingual sentence pairs along with corresponding term alignments.
For each test example, we input only the source sentence into the model and assess whether the expected target term appears correctly in the generated translation. We report terminology accuracy—the proportion of reference target terms correctly generated in the translation—in both Zh$\Rightarrow$En and De$\Rightarrow$En directions, as shown in Table~\ref{tab:termacc}.

In the Zh$\Rightarrow$En setting, LRMs demonstrate superior performance, with DeepSeek-R1 achieving the highest accuracy (53.20\%).
We attribute this to LRMs’ stronger reasoning and contextual comprehension capabilities, which allow them to better resolve semantic ambiguity in Chinese and plan conceptual mappings into English, ensuring the naturalness and accuracy of terminology in the English context.
These results suggest that LRMs hold strong potential in terminology translation.
In contrast, traditional LLMs outperform LRMs in the De$\Rightarrow$En direction, with Gemini-2.0-Flash achieving the highest accuracy (38.55\%).
This may be attributed to the high lexical similarity between German and English, which allows for more direct term mapping, favoring models trained to mimic reference terms without requiring deep reasoning.

Therefore, in practical applications, model selection should take into account the linguistic gap between source and target languages: LRMs are better suited for language pairs with significant structural differences (\emph{e.g.}, Zh$\Rightarrow$En), while traditional LLMs may be more effective for structurally similar pairs like De$\Rightarrow$En.

Given these findings, we argue that improving the terminology translation performance of LRMs requires tighter integration between their reasoning structures and terminology modeling capabilities. Future work may explore embedding terminology constraints into intermediate reasoning chains to support explicit term-level planning. These structural and reasoning-aligned improvements are expected to unlock the potential of LRMs in terminology-intensive translation scenarios.

\begin{table*}[t]
\centering
\resizebox{\textwidth}{!}{%
\begin{tabular}{l ccc ccc ccc ccc ccc}
\toprule
\multirow{2}{*}{Model} & \multicolumn{3}{c}{Conversation} & \multicolumn{3}{c}{E-commerce} & \multicolumn{3}{c}{News} & \multicolumn{3}{c}{Social} & \multicolumn{3}{c}{Literary} \\
\cmidrule(lr){2-4} \cmidrule(lr){5-7} \cmidrule(lr){8-10} \cmidrule(lr){11-13} \cmidrule(lr){14-16}
 & BLEU & COMET & KIWI & BLEU & COMET & KIWI & BLEU & COMET & KIWI & BLEU & COMET & KIWI & BLEU & COMET & KIWI \\
\midrule
\multicolumn{16}{c}{\textbf{Document-Level}} \\
GPT-4o & \textbf{40.00} & 84.71 & 82.02 & 35.33 & 87.35 & 64.55 & 36.07 & 86.60 & 66.01 & 33.75 & 84.36 & 68.70 & 17.55 & \textbf{77.57} & 77.14 \\
DeepSeek-V3 & 39.28 & 84.51 & 81.98 & 36.71 & 87.26 & 63.16 & 37.59 & 86.50 & 65.35 & 36.48 & 84.80 & 67.83 & 16.81 & 77.48 & 77.17 \\
Gemini-2.0-Flash & 39.98 & 84.64 & 81.81 & \textbf{37.72} & 87.34 & 62.33 & 38.68 & 85.71 & 63.68 & \textbf{37.76} & 84.38 & 67.14 & \textbf{18.12} & 77.30 & 76.62 \\
\hdashline
OpenAI-o1 & 37.05 & \textbf{84.83} & \textbf{82.99} & 30.68 & \textbf{87.40} & \textbf{66.50} & 32.08 & \textbf{86.67} & \textbf{67.90} & 29.02 & \textbf{84.82} & \textbf{69.79} & 16.44 & \textbf{77.57} & \textbf{77.76} \\
DeepSeek-R1 & 33.66 & 84.15 & 81.45 & 21.62 & 85.23 & 63.77 & 23.49 & 83.68 & 66.99 & 22.22 & 81.59 & 68.09 & 13.12 & 76.64 & 76.47 \\
OpenAI-o3-mini & 36.58 & 84.52 & 82.05 & 32.13 & 87.15 & 65.18 & 33.60 & 86.22 & 65.77 & 30.93 & 84.76 & 68.87 & 17.26 & 76.87 & 76.64 \\
Gemini-2.0-Flash-Thinking & \textbf{40.00} & 84.59 & 81.79 & 37.41 & 87.35 & 62.39 & \textbf{38.82} & 86.62 & 64.19 & 36.97 & 84.55 & 66.47 & 17.77 & 77.17 & 76.58 \\
\midrule
\midrule
\multicolumn{16}{c}{\textbf{Sentence-Level}} \\
GPT-4o & \textbf{38.68} & 87.56 & 82.00 & \textbf{34.17} & 86.38 & 81.78 & 36.21 & 86.05 & 82.66 & 33.07 & 84.63 & 81.47 & 18.53 & \textbf{82.96} & \textbf{57.23} \\
DeepSeek-V3 & 37.46 & 87.61 & 81.99 & 33.57 & 86.32 & 81.93 & 35.82 & 86.30 & 83.04 & 32.96 & 84.74 & 81.81 & 19.92 & 82.88 & 54.68 \\
Gemini-2.0-Flash & 38.01 & 87.29 & 81.87 & 33.67 & 86.36 & 81.67 & \textbf{36.48} & 86.21 & 82.80 & \textbf{33.24} & 84.55 & 81.49 & 23.28 & 82.80 & 53.83 \\
\hdashline
OpenAI-o1 & 36.30 & \textbf{87.87} & \textbf{83.49} & 31.30 & \textbf{86.39} & \textbf{83.05} & 33.17 & \textbf{86.35} & \textbf{84.35} & 29.95 & \textbf{84.80} & \textbf{82.96} & 16.87 & 82.42 & 55.73 \\
DeepSeek-R1 & 34.05 & 87.13 & 81.93 & 29.23 & 85.95 & 81.75 & 31.83 & 86.03 & 83.01 & 28.51 & 84.26 & 81.59 & 5.16 & 80.35 & 52.16 \\
OpenAI-o3-mini & 36.33 & 86.74 & 81.67 & 32.81 & 85.92 & 81.73 & 33.95 & 85.57 & 82.43 & 31.88 & 84.07 & 81.28 & 18.06 & 82.83 & 55.16 \\
Gemini-2.0-Flash-Thinking & 37.07 & 86.99 & 81.68 & 33.09 & 86.17 & 81.55 & 35.70 & 86.19 & 82.81 & 32.30 & 84.16 & 81.27 & \textbf{23.52} & 82.77 & 53.69 \\
\bottomrule
\end{tabular}%
}
\caption{Average BLEU, COMET, and CometKiwi scores at document-level and sentence-level across five domains.}
\label{tab:length_analysis}
\end{table*}

%length
\subsection{Document-Level Evaluation}
To investigate the impact of textual granularity on model performance, we compare sentence-level and document-level inputs across five domains, as shown in Table~\ref{tab:length_analysis}. 
Compared to traditional LLMs, LRMs demonstrate stronger adaptability in document-level translation. They consistently achieve higher scores on COMET and CometKiwi, which focus on semantic similarity and contextual coherence. This advantage is particularly evident in long-text scenarios, where modeling inter-sentential dependencies and broader discourse structures is essential. Recent studies have shown that LRMs leverage Long CoT reasoning to enhance translation quality in complex texts, effectively modeling longer-range dependencies and maintaining coherence across extended contexts~\cite{liu2025newtrendsmodernmachine}. In domains like Conversation and News, which naturally involve multi-turn interactions and cross-sentence semantic dependencies, LRMs are better equipped to integrate information across spans and plan discourse globally, enabling both lexical and semantic alignment in context-sensitive scenarios.

These findings highlight the advantages of LRMs for document-level translation. Their capacity for planning, reasoning, and generating globally coherent outputs makes them better suited for long-form and context-sensitive scenarios.
Future research could further explore the adaptive translation capabilities of LRMs across different document types, such as legal contracts, technical manuals, and literary works.

\subsection{Effect of Domain-Aware Prompt}

To further investigate the underperformance of LRMs in domain-intensive settings such as Law, Medical, and IT, we hypothesize that these models may lack sufficient domain sensitivity during translation reasoning, potentially due to the limitations imposed by the prompts in accurately guiding the model to recognize and adapt to domain-specific characteristics. Therefore, we experiment with three prompting strategies (Table~\ref{tab:prompt_templates}): P1, where no domain-specific information is provided; P2, where explicit domain labels are given; and P3, where the model is encouraged to infer the domain and adjust its translation style accordingly.

Using DeepSeek-R1 as a representative model, we evaluate the three prompting strategies across IT, Law, and Medical domains. As shown in Table~\ref{tab:prompt_results}, P3 achieves the highest COMET and CometKiwi scores across all domains, indicating its effectiveness in producing semantically faithful and stylistically coherent translations. 
Both P2 and P3 significantly outperform P1 on semantic metrics, indicating that incorporating domain awareness—either explicitly or implicitly—substantially improves contextual understanding and semantic coherence.
While P1 and P3 yield similar BLEU scores in the IT domain, P3 substantially outperforms P1 in Law and Medical, especially in CometKiwi, suggesting that domain-aware reasoning contributes to improved contextual understanding. 
Interestingly, P2 achieves the highest BLEU scores in the Law domain, likely likely because its explicit domain cues align closely with the rigid structure and formulaic phrasing typical of legal texts. 
However, P2 lags behind P3 on COMET and CometKiwi, suggesting that fixed labels may limit contextual flexibility. 
In contrast, P3 allows the model to infer domain style from context, resulting in more fluent and semantically coherent translations.

These findings highlight the superiority of domain-adaptive prompting over static label specification in MDMT. Future work may focus on injecting domain inference into the reasoning chain, enabling LRMs to autonomously determine the domain and adjust their translation accordingly.

\begin{table}[t]
\centering
\resizebox{.48\textwidth}{!}{%
\begin{tabular}{@{}ll@{}}
\toprule
\textbf{ID} & \textbf{Prompt} \\
\midrule
\textbf{P1} & Translate the following \texttt{\{src\_lang\}} sentence into \texttt{\{tgt\_lang\}}. \\
\textbf{P2} & Given the \texttt{\{src\_lang\}} sentence from a \texttt{\{domain\}}-related text, \\
           & translate it into \texttt{\{tgt\_lang\}} in a \texttt{\{domain\}} domain style. \\
\textbf{P3} & Translate the following \texttt{\{src\_lang\}} text into \texttt{\{tgt\_lang\}} while \\
            & maintaining the domain style of the source text. \\
\bottomrule
\end{tabular}
}
\caption{Prompts used in domain-aware translation.}
\label{tab:prompt_templates}
\end{table}

\begin{table}[t]
\centering
\resizebox{\linewidth}{!}{%
\begin{tabular}{l ccc ccc ccc}
\toprule
\multirow{2}{*}{Prompt} & \multicolumn{3}{c}{IT} & \multicolumn{3}{c}{Law} & \multicolumn{3}{c}{Medical} \\
\cmidrule(lr){2-4} \cmidrule(lr){5-7} \cmidrule(lr){8-10}
& BLEU & COMET & KIWI & BLEU & COMET & KIWI & BLEU & COMET & KIWI \\
\midrule
P1 & \textbf{36.66} & 83.49 & 78.68 & 39.16 & 84.97 & 82.35 & 40.23 & 83.82 & 81.81 \\
P2 & 34.20 & 82.94 & 79.33 & \textbf{41.46} & 85.01 & 83.46 & 38.92 & 83.38 & 82.57 \\
P3 & 36.55 & \textbf{83.57} & \textbf{79.82} & 40.57 & \textbf{85.32} & \textbf{83.54} & \textbf{41.14} & \textbf{84.06} & \textbf{82.97} \\
\bottomrule
\end{tabular}
}
\caption{Translation performance across IT, Law, and Medical domains under different prompt strategies.}
\label{tab:prompt_results}
\end{table}

\begin{table*}[t]
\centering
\resizebox{\textwidth}{!}{%
\begin{tabular}{l ccc ccc ccc ccc ccc}
\toprule
\multirow{2}{*}{Model} & \multicolumn{3}{c}{Level 1} & \multicolumn{3}{c}{Level 2} & \multicolumn{3}{c}{Level 3} & \multicolumn{3}{c}{Level 4} & \multicolumn{3}{c}{Level 5} \\
\cmidrule(lr){2-4} \cmidrule(lr){5-7} \cmidrule(lr){8-10} \cmidrule(lr){11-13} \cmidrule(lr){14-16}
 & BLEU & COMET & KIWI & BLEU & COMET & KIWI & BLEU & COMET & KIWI & BLEU & COMET & KIWI & BLEU & COMET & KIWI \\
\midrule
GPT-4o & \textbf{41.96} & \textbf{85.13} & 79.80 & 30.08 & 84.88 & 82.15 & 26.49 & 82.75 & 80.91 & 22.61 & 83.11 & 75.48 & 24.94 & 83.59 & 67.11 \\
DeepSeek-V3 & 40.81 & 85.09 & 79.90 & 27.90 & 84.86 & 82.20 & 25.43 & 82.54 & 80.97 & 22.57 & 82.95 & 75.30 & 25.06 & 83.30 & 67.64 \\
Gemini-2.0-Flash & 41.15 & 84.39 & 79.46 & 29.22 & 84.93 & 81.96 & 27.53 & 82.80 & 80.68 & 24.53 & 82.72 & 74.57 & 28.93 & 83.91 & 66.28 \\
\hdashline
OpenAI-o1 & 40.34 & 84.59 & \textbf{81.10} & 29.67 & 84.88 & \textbf{83.28} & 25.76 & \textbf{82.94} & \textbf{81.94} & 21.99 & \textbf{83.31} & \textbf{76.92} & 22.78 & \textbf{84.02} & \textbf{69.62} \\
DeepSeek-R1 & 41.43 & 84.33 & 79.86 & 27.12 & 84.25 & 81.75 & 24.32 & 82.15 & 80.44 & 18.66 & 82.17 & 75.27 & 19.53 & 82.43 & 68.78 \\
OpenAI-o3-mini & 40.74 & 84.12 & 79.44 & 26.49 & 84.56 & 82.14 & 24.93 & 82.36 & 80.73 & 21.09 & 82.67 & 75.37 & 22.70 & 83.27 & 67.63 \\
Gemini-2.0-Flash-Thinking & 39.87 & 83.84 & 79.36 & \textbf{30.78} & \textbf{84.98} & 82.21 & \textbf{28.15} & 82.68 & 80.62 & \textbf{24.54} & 83.02 & 74.71 & \textbf{28.97} & 83.93 & 66.32 \\
\bottomrule
\end{tabular}%
}
\caption{Performance across different levels of complexity, comparing LRMs and tranditional LLMs.}
\label{tab:metrics_by_level_avg}
\end{table*}

% difficulty
\subsection{Impact of Translation Difficulty}
In recent years, researchers have proposed various reasoning-enhanced strategies to improve the performance of LLMs on complex tasks~\cite{wei2022chain, yao2023tree, zhou2022least}. These approaches focus on guiding models through structured intermediate reasoning to better capture semantic and logical complexity. When extended to the field of MT, we hypothesize that incorporating mechanisms such as problem decomposition, reasoning path exploration, and stepwise inference can enhance a model’s generalization ability in tasks with high semantic load. The LRMs exemplify this reasoning-oriented design philosophy. These models are trained with structured cognitive capabilities, which allows them to operate more robustly in linguistically complex translation scenarios.

To systematically assess the impact of translation difficulty on model performance, we use DeepSeek-V3 to rate the complexity of source sentences drawn from three datasets: Multi-Domain (De$\Rightarrow$En), WMT22 (De$\Leftrightarrow$En and Zh$\Leftrightarrow$En), and Guofeng WebNovel (Zh$\Rightarrow$En). These sentences are classified into five difficulty levels, ranging from Level 1, representing the easiest, to Level 5, representing the most difficult. 
The prompt used for difficulty assessment is provided in Figure~\ref{fig:difficulty_eval}. 
As shown in Table~\ref{tab:metrics_by_level_avg}, traditional LLMs perform well on simpler inputs, especially at Level 1 where they achieve high BLEU and COMET scores. 
However, their performance declines significantly as difficulty increases. Starting at Level 2, they are consistently outperformed by LRMs. 
In particular, OpenAI-o1 and Gemini-2.0-Flash-Thinking demonstrate stronger robustness and semantic consistency under higher difficulty conditions, maintaining clear advantages on metrics such as COMET and CometKiwi.
This trend supports our hypothesis that LRMs are better equipped to model and generalize in semantically dense and contextually demanding translation tasks.

Future research should further investigate how LRMs adapt to varying levels of translation difficulty.
For example, LRMs should generate accurate translations efficiently with minimal reasoning for simple inputs to avoid unnecessary overthinking~\cite{chen2025overthinking}, whereas for more complex sentences, they should engage in deeper analysis and structured reasoning to ensure semantic accuracy and completeness.

% token Consumption Comparison
\subsection{Token Consumption Comparison}
We compare the average token usage of traditional LLMs and LRMs on the UM-Corpus (En$\Rightarrow$Zh) multi-domain test sets to assess translation efficiency. As shown in Table~\ref{tab:um_results_averages}, LRMs generate significantly longer outputs than traditional LLMs. While they achieve strong CometKiwi scores, their lower performance on COMET and BLEU suggests that LRMs may "overthink" during generation~\cite{chen2025overthinking}. In MDMT, future work could explore adapting reasoning depth to translation complexity to reduce overthinking.
\begin{table}[t]
\centering
\resizebox{\linewidth}{!}{%
\begin{tabular}{l c c c c}
\toprule
\textbf{Model} & \textbf{BLEU} & \textbf{COMET} & \textbf{KIWI} & \textbf{Avg. Len.} \\
\midrule
GPT-4o & 37.40 & 86.74 & 84.61 & 31.76 \\
DeepSeek-V3 & \textbf{43.34} & \textbf{87.76} & 85.04 & \textbf{29.63} \\
Gemini-2.0-Flash & 40.30 & 87.05 & 84.80 & 32.71 \\
\hdashline
OpenAI-o1 & 36.17 & 87.11 & \textbf{85.09} & 402.64 \\
DeepSeek-R1 & 41.12 & 87.50 & 85.03 & 485.05 \\
OpenAI-o3-mini & 34.60 & 86.69 & 84.98 & 424.31 \\
Gemini-2.0-Flash-Thinking & 39.17 & 86.63 & 84.37 & 1043.59 \\
\bottomrule
\end{tabular}%
}
\caption{Average BLEU, COMET, KIWI scores, and response length for various LLMs.}
\label{tab:um_results_averages}
\end{table}

% case study
\subsection{Case Study}
We present a case from the \textit{lexical\_ambiguity} subset of the CommonMT (Zh$\Rightarrow$En) test set. The source sentence contains a figurative expression referring to the act of leaving a stable job to pursue private business opportunities. DeepSeek-R1 accurately captures this meaning and generates: \textit{“In the 1990s, many people ventured into business,”} which matches the reference exactly. Compared to alternatives such as “went into business,” R1’s choice of “ventured” better reflects the intended nuance, demonstrating its ability to resolve lexical ambiguity through contextual reasoning.

\begin{figure}[t]
    \centering
    \includegraphics[width=\linewidth]{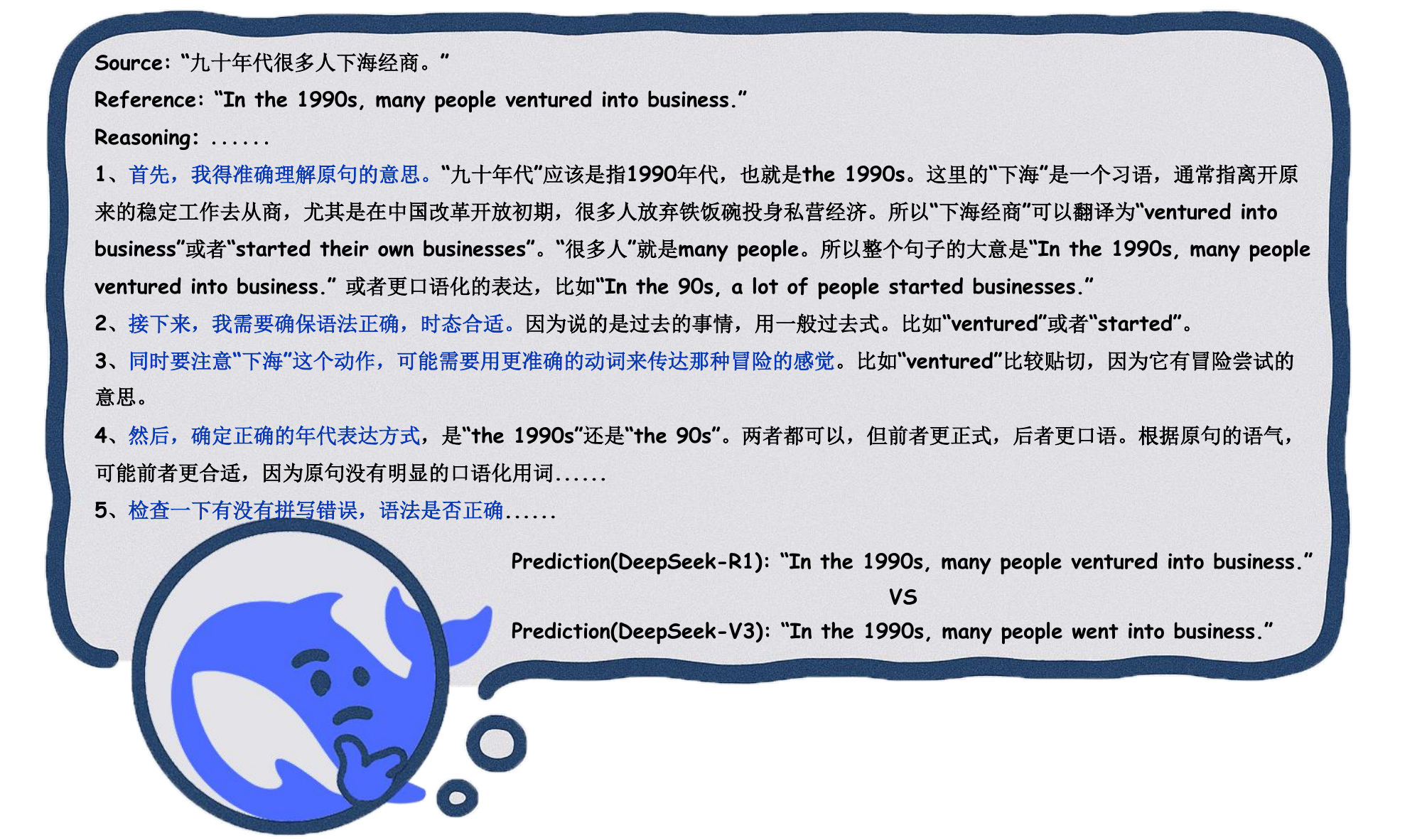}
    \caption{Case study of DeepSeek-R1 showing its reasoning process and final translation. }
    \label{fig:case_study}
\end{figure}

\section{Conclusion}

This paper presents a comprehensive evaluation of LRMs in MDMT. Through systematic comparison with traditional LLMs across 15 domains and four translation directions, we observe that LRMs achieve superior performance in semantically challenging and long-form translation tasks, as reflected in meaning-oriented metrics such as COMET and CometKiwi. However, their performance declines in terminology-intensive and stylistically constrained domains, where traditional LLMs exhibit stronger lexical precision and structural fidelity. Our prompt-based analysis further demonstrates the advantage of domain-adaptive prompting in improving contextual alignment. These findings suggest that future work should focus on integrating structured reasoning with domain-specific constraints to enhance the reliability and controllability of LRMs in real-world translation scenarios.

\section*{Limitations}

Despite providing a comprehensive evaluation of LRMs in MDMT, our study has several limitations. First, the evaluation is restricted to high-resource language pairs, which may limit the generalizability of our findings to low-resource or typologically distant languages. Second, our automatic MQM annotations rely on a single model as the scorer, which ensures consistency but may introduce bias aligned with that model’s own translation behavior. 

\bibliography{acl_latex}

\appendix
\newpage

\section{Details of Multi-domain Test Sets}
\label{appendix:dataset}
We compile a collection of publicly available translation test sets covering diverse domains and four translation directions: De$\Rightarrow$En, En$\Rightarrow$De, En$\Rightarrow$Zh, and Zh$\Rightarrow$En. The datasets span general-purpose and domain-specific content, including biomedical, legal, IT, conversational, and literary texts, etc. Sample sizes for each domain are reported in Tables~\ref{tab:test_zh-en} to~\ref{tab:test_en-de}, with no additional filtering or augmentation applied.

\begin{table}[t]
\centering
\small
% \resizebox{.5\textwidth}{!}{%
\begin{tabular}{lc lc}
\toprule
\textbf{Test set} & \textbf{Num} & \textbf{Test set} & \textbf{Num} \\
\midrule
Conversation & 349 & Social & 491 \\
E-commerce & 518 & News (WMT22) & 499 \\
Bio (WMT18) & 212 & Bio (WMT19) & 243 \\
Commonsense & 1200 & Literary & 3038 \\
Terminology & 1638 & Mixed (WMT22) & 1875 \\
\bottomrule
\end{tabular}%
% }
\caption{Test sets and the number of samples for Zh$\Rightarrow$En translation tasks. Bio denotes the Biomedical domain.}
\label{tab:test_zh-en}
\end{table}

\begin{table}[t]
\centering
\small
% \resizebox{.5\textwidth}{!}{%
\begin{tabular}{lc lc}
\toprule
\textbf{Test set} & \textbf{Num} & \textbf{Test set} & \textbf{Num} \\
\midrule
Culture & 778 & Laws & 456 \\
Subtitles & 597 & Science & 503 \\
Conversation & 484 & Social & 511 \\
E-commerce & 530 & News (WMT22) & 511 \\
Bio (WMT18) & 244 & Bio (WMT19) & 224 \\
News (UM) & 1500 & Mixed & 2037 \\
\bottomrule
\end{tabular}%
% }
\caption{Test sets and the number of samples for En$\Rightarrow$Zh translation tasks.}
\label{tab:test_en-zh}
\end{table}

\begin{table}[t]
\centering
\small
% \resizebox{.5\textwidth}{!}{%
\begin{tabular}{lc lc}
\toprule
\textbf{Test set} & \textbf{Num} & \textbf{Test set} & \textbf{Num} \\
\midrule
Bio (WMT18) & 217 & Bio (WMT19) & 373 \\
Social & 515 & E-commerce & 501 \\
News (WMT22) & 506 & Conversation & 462  \\
IT & 2000 & Law & 2000 \\
Koran & 2000 & Subtitles & 2000 \\
Medical & 2000 & Terminology & 2955 \\
Mixed & 1984 & & \\
\bottomrule
\end{tabular}%
% }
\caption{Test sets and the number of samples for De$\Rightarrow$En translation tasks.}
\label{tab:test_de-en}
\end{table}

\begin{table}[t]
\centering
\small
% \resizebox{.5\textwidth}{!}{%
\begin{tabular}{lc lc}
\toprule
\textbf{Test set} & \textbf{Num} & \textbf{Test set} & \textbf{Num} \\
\midrule
Bio (WMT18) & 220 & Bio (WMT19) & 435 \\
News (WMT22) & 511 & E-commerce & 530 \\
Social & 512 & Conversation & 484 \\
Mixed & 2037 & & \\
\bottomrule
\end{tabular}%
% }
\caption{Test sets and the number of samples for En$\Rightarrow$De translation tasks.}
\label{tab:test_en-de}
\end{table}

\section{MQM}
\label{appendix:mqm}
To evaluate translation quality across domains and systems, we adopt an enhanced MQM hierarchy tailored to the characteristics of LLMs. Based on the official MQM taxonomy, we introduce additional error types frequently observed in LLM outputs—such as hallucination, semantic repetition, and cross-domain confusion. Our final schema spans seven dimensions: Accuracy, Fluency, Style, Terminology, Others, Source Error, and Non-translation Error, covering 25 fine-grained error categories (Table~\ref{tab:mqm-hierarchy}).

For consistent and scalable annotation, we employ DeepSeek-V3 as the automatic scoring model across all evaluations. It is applied uniformly to both traditional and LRMs to ensure fair comparison. The model follows a structured prompt designed to mimic human assessment while enforcing strict formatting and error attribution rules. The full prompt is shown in Figure~\ref{fig:mqm-prompt}.

\begin{table*} [t]
  \centering
  \small
  \begin{tabularx}{\textwidth}{l l X}
    \toprule
    \textbf{Category} & \textbf{Error Type} & \textbf{Description} \\
    \midrule
    \multirow{8}{*}{Accuracy} 
      & Mistranslation & Inaccurate translation causing semantic distortion. \\
      & Addition & Adding extra information or emotions not in the source.\\
      & Under-translation & Failure to fully convey cultural or contextual nuances. \\
      & Omission & Unintentional exclusion of content from the source text. \\
      & Untranslated & Retaining source text without translation. \\
      & Hallucination & Generating content unrelated to the source text. \\
      & Off-target Translation & Translation misalignment caused by ambiguous input. \\  
      & Contradiction & Contradicting itself or the source text. \\  
    \midrule
    \multirow{5}{*}{Fluency} 
      & Grammar & Errors in sentence structure or syntax. \\
      & Punctuation & Incorrect use of punctuation marks. \\
      & Spelling & Misspelling of words. \\
      & Semantic Repetition & Unnecessary repetition of words or phrases. \\
      & Logical Incoherence & Lack of logical flow or coherence in translation. \\
    \midrule
    \multirow{4}{*}{Style} 
      & Awkward Expression & Stilted or unnatural phrasing in the target language. \\
      & Unidiomatic Usage & Literal translation causes unnatural wording or grammar. \\
      & Style Inconsistency & Inconsistent stylistic choices within the translation. \\
      & Over-localization & Excessive cultural adaptation leading to distortion. \\
    \midrule
    \multirow{5}{*}{Terminology} 
      & Terminology Inconsistency & Inconsistent translation of the same term. \\
      & Terminology Misuse & Use of incorrect or inappropriate terms for the domain. \\
      & Cross-domain Confusion & Misuse of terms due to domain shifts. \\
      & Incorrect Unit Conversion & Errors in unit conversion (e.g., metric to imperial). \\  
      & Formatting & Errors in domain-specific formats (e.g., legal, medical). \\  
    \midrule
    \multirow{1}{*}{Others} 
      & & Any other errors not covered in the above categories. \\
    \midrule
    \multirow{1}{*}{Source Error} 
      & & Errors present in the Source text itself. \\
    \midrule
    \multirow{1}{*}{Non-translation Error} 
      & & Translation is unassessable and unrelated to the Source. \\
    \bottomrule
  \end{tabularx}
  \caption{\label{tab:mqm-hierarchy} MQM Hierarchy.}
\end{table*}

\section{Evaluation Results}

We report detailed automatic evaluation results across all domains and translation directions using BLEU, COMET, and CometKiwi. These results are summarized in Tables~\ref{tab:result_culture_mt} through~\ref{tab:result_wmt22}, covering all test sets introduced in our main experiments. Each table corresponds to a specific dataset or language direction, facilitating fine-grained comparison across models.

\begin{table}[t]
\centering
\resizebox{\linewidth}{!}{%
\begin{tabular}{l ccc}
\toprule
\multirow{2}{*}{Model} & \multicolumn{3}{c}{Culture} \\
\cmidrule(lr){2-4}
 & BLEU & COMET & KIWI \\
\midrule
GPT-4o & 38.65 & 85.24 & 83.14 \\
DeepSeek-V3 & \textbf{40.27} & \textbf{85.46} & 83.21 \\
Gemini-2.0-Flash & 39.11 & 85.02 & 82.94 \\
\hdashline
OpenAI-o1 & 34.57 & 85.06 & \textbf{84.25} \\
DeepSeek-R1 & 35.88 & 85.45 & 83.14 \\
OpenAI-o3-mini & 34.05 & 84.36 & \textbf{84.25} \\
Gemini-2.0-Flash-Thinking & 36.94 & 84.63 & 83.68 \\
\bottomrule
\end{tabular}
}
\caption{Results on the En$\Rightarrow$Zh Culture-MT dataset.}
\label{tab:result_culture_mt}
\end{table}

\begin{table*}[h]
\centering
\resizebox{\textwidth}{!}{%
\begin{tabular}{lccc ccc ccc ccc}
\toprule
\multirow{2}{*}{Model} 
& \multicolumn{3}{c}{Contextual Syntactic} 
& \multicolumn{3}{c}{Contextless Syntactic} 
& \multicolumn{3}{c}{Lexical} 
& \multicolumn{3}{c}{Average} \\
\cmidrule(lr){2-4} \cmidrule(lr){5-7} \cmidrule(lr){8-10} \cmidrule(lr){11-13}
& BLEU & COMET & KIWI & BLEU & COMET & KIWI & BLEU & COMET & KIWI & BLEU & COMET & KIWI \\
\midrule
GPT-4o & 31.42 & \textbf{85.11} & 81.88 & 32.82 & 85.24 & 78.35 & 33.89 & 85.41 & 79.83 & 32.71 & 85.25 & 80.02 \\
DeepSeek-V3 & 30.19 & 84.90 & 81.91 & \textbf{33.53} & \textbf{85.58} & 78.76 & \textbf{35.09} & \textbf{85.52} & 79.69 & \textbf{32.94} & \textbf{85.33} & 80.12 \\
Gemini-2.0-Flash & 31.15 & 84.86 & 81.79 & 31.69 & 85.20 & 77.99 & 32.13 & 84.03 & 78.84 & 31.66 & 84.70 & 79.54 \\
\hdashline
OpenAI-o1 & 29.80 & 84.74 & 80.66 & 28.04 & 85.14 & 79.48 & 30.40 & 84.00 & 79.70 & 28.50 & 84.96 & \textbf{81.11} \\
DeepSeek-R1 & 26.28 & 84.54 & \textbf{82.07} & 30.00 & 84.96 & 78.62 & 30.40 & 84.00 & 79.70 & 28.89 & 84.50 & 80.13 \\
OpenAI-o3-mini & 28.55 & 84.57 & 79.19 & 27.18 & 84.43 & \textbf{82.97} & 29.43 & 83.82 & \textbf{80.87} & 28.39 & 84.27 & 81.01 \\
Gemini-2.0-Flash-Thinking & \textbf{31.54} & 84.61 & 81.96 & 31.60 & 84.34 & 77.66 & 33.49 & 84.59 & 79.95 & 32.21 & 84.51 & 79.86 \\
\bottomrule
\end{tabular}
}
\caption{Results on the Zh$\Rightarrow$En Commonsense-MT dataset.}
\label{tab:zh-en-ambiguity}
\end{table*}

\begin{table*}[h]
\centering
\resizebox{\textwidth}{!}{%
\begin{tabular}{l ccc ccc ccc ccc}
\toprule
\multirow{2}{*}{Model} & \multicolumn{3}{c}{News} & \multicolumn{3}{c}{Laws} & \multicolumn{3}{c}{Science} & \multicolumn{3}{c}{Subtitles} \\
\cmidrule(lr){2-4} \cmidrule(lr){5-7} \cmidrule(lr){8-10} \cmidrule(lr){11-13}
 & BLEU & COMET & KIWI & BLEU & COMET & KIWI & BLEU & COMET & KIWI & BLEU & COMET & KIWI \\
\midrule
GPT-4o & 34.58 & 86.22 & 85.02 & 58.00 & 91.50 & 86.65 & 32.09 & 87.34 & 84.84 & \textbf{24.91} & 81.91 & 81.93 \\
DeepSeek-V3 & 35.86 & \textbf{86.94} & 85.44 & \textbf{80.33} & \textbf{93.36} & 86.55 & \textbf{33.54} & \textbf{88.28} & 85.46 & 23.66 & \textbf{82.44} & 82.69 \\
Gemini-2.0-Flash & \textbf{36.39} & 86.37 & 85.14 & 66.71 & 92.23 & 86.49 & 33.47 & 87.69 & 85.24 & 24.64 & 81.91 & 82.33 \\
\hdashline
OpenAI-o1 & 33.93 & 86.61 & \textbf{85.47} & 55.48 & 91.61 & 86.73 & 32.32 & 87.92 & \textbf{85.56} & 22.96 & 82.31 & 82.60 \\
DeepSeek-R1 & 33.35 & 86.67 & 85.40 & 78.81 & 93.16 & 86.40 & 31.82 & 88.08 & 85.47 & 20.50 & 82.10 & \textbf{82.84} \\
OpenAI-o3-mini & 32.90 & 86.29 & 85.35 & 51.18 & 91.15 & \textbf{86.92} & 31.38 & 87.42 & 85.24 & 22.95 & 81.90 & 82.40 \\
Gemini-2.0-Flash-Thinking & 33.90 & 86.02 & 84.79 & 66.99 & 92.12 & 86.40 & 31.98 & 87.47 & 85.06 & 23.83 & 80.90 & 81.22 \\
\bottomrule
\end{tabular}%
}
\caption{Results on the En$\Rightarrow$Zh UM-Corpus dataset.}
\label{tab:result_um_domains}
\end{table*}

\begin{table*}[h]
\centering
\resizebox{\textwidth}{!}{%
\begin{tabular}{l ccc ccc ccc ccc ccc}
\toprule
\multirow{2}{*}{Model} & \multicolumn{3}{c}{Conversation} & \multicolumn{3}{c}{E-commerce} & \multicolumn{3}{c}{News} & \multicolumn{3}{c}{Social} & \multicolumn{3}{c}{Average} \\
\cmidrule(lr){2-4} \cmidrule(lr){5-7} \cmidrule(lr){8-10} \cmidrule(lr){11-13} \cmidrule(lr){14-16}
 & BLEU & COMET & KIWI & BLEU & COMET & KIWI & BLEU & COMET & KIWI & BLEU & COMET & KIWI & BLEU & COMET & KIWI \\
\midrule
GPT-4o & 31.28 & 84.86 & 79.83 & 19.94 & 81.75 & 78.24 & 25.21 & \textbf{83.04} & 81.10 & 27.29 & \textbf{83.77} & 80.64 & 25.93 & \textbf{83.36} & 79.95 \\
DeepSeek-V3 & 30.26 & \textbf{84.96} & 79.67 & 19.47 & 81.55 & 78.32 & 25.77 & 83.00 & 81.18 & 27.55 & 83.56 & 80.64 & 25.76 & 83.27 & 79.95 \\
Gemini-2.0-Flash & \textbf{32.79} & 84.46 & 79.19 & \textbf{19.97} & 81.54 & 77.90 & \textbf{26.62} & \textbf{83.04} & 80.93 & \textbf{29.25} & 83.24 & 80.16 & \textbf{27.16} & 83.07 & 79.55 \\
\hdashline
OpenAI-o1 & 29.52 & 84.95 & \textbf{80.90} & 17.82 & \textbf{81.81} & 79.16 & 22.52 & 82.96 & \textbf{82.10} & 23.85 & 83.20 & \textbf{81.39} & 23.43 & 83.23 & \textbf{80.89} \\
DeepSeek-R1 & 26.82 & 84.61 & 79.40 & 15.63 & 80.98 & 77.87 & 21.21 & 82.34 & 80.95 & 22.18 & 82.69 & 80.05 & 21.46 & 82.66 & 79.57 \\
OpenAI-o3-mini & 30.57 & 84.25 & 79.48 & 18.80 & 81.29 & 78.23 & 23.68 & 82.88 & 81.16 & 26.98 & 83.34 & 80.61 & 25.01 & 82.94 & 79.87 \\
Gemini-2.0-Flash-Thinking & 30.98 & 84.17 & 79.28 & 19.06 & 81.21 & 77.76 & 25.50 & 82.90 & 80.85 & 27.54 & 83.04 & 80.13 & 25.77 & 82.83 & 79.50 \\
\bottomrule
\end{tabular}%
}
\caption{Results on the Zh$\Rightarrow$En WMT22 General Machine Translation task.}
\end{table*}

\begin{table*}[h]
\centering
\resizebox{\textwidth}{!}{%
\begin{tabular}{l ccc ccc ccc ccc ccc}
\toprule
\multirow{2}{*}{Model} & \multicolumn{3}{c}{Conversation} & \multicolumn{3}{c}{E-commerce} & \multicolumn{3}{c}{News} & \multicolumn{3}{c}{Social} & \multicolumn{3}{c}{Average} \\
\cmidrule(lr){2-4} \cmidrule(lr){5-7} \cmidrule(lr){8-10} \cmidrule(lr){11-13} \cmidrule(lr){14-16}
 & BLEU & COMET & KIWI & BLEU & COMET & KIWI & BLEU & COMET & KIWI & BLEU & COMET & KIWI & BLEU & COMET & KIWI \\
\midrule
GPT-4o & \textbf{45.63} & 89.44 & 82.67 & \textbf{43.82} & 88.82 & 83.52 & 51.20 & 88.31 & 83.41 & \textbf{38.60} & 84.57 & 81.33 & \textbf{44.81} & 87.78 & 82.73 \\
DeepSeek-V3 & 43.49 & 89.89 & 82.94 & 42.84 & 89.00 & 83.79 & 50.28 & \textbf{88.75} & 83.76 & 37.59 & \textbf{85.27} & 82.24 & 43.55 & \textbf{88.23} & 83.18 \\
Gemini-2.0-Flash & 44.59 & \textbf{90.19} & 83.01 & 42.95 & 88.90 & 83.47 & \textbf{52.12} & 88.46 & 83.28 & 38.11 & 84.71 & 81.59 & 44.44 & 88.06 & 82.84 \\
\hdashline
OpenAI-o1 & 42.24 & 89.86 & \textbf{84.13} & 39.22 & \textbf{89.06} & \textbf{84.59} & 45.88 & 88.47 & \textbf{84.67} & 33.41 & 85.07 & \textbf{82.98} & 40.19 & 88.11 & \textbf{84.09} \\
DeepSeek-R1 & 38.00 & 88.74 & 82.74 & 34.48 & 88.35 & 83.48 & 42.56 & 88.08 & 83.58 & 29.43 & 84.21 & 82.01 & 36.12 & 87.34 & 82.95 \\
OpenAI-o3-mini & 43.24 & 89.02 & 82.44 & 41.54 & 88.51 & 83.41 & 47.06 & 87.53 & 82.94 & 36.38 & 83.99 & 81.27 & 42.05 & 87.26 & 82.52 \\
Gemini-2.0-Flash-Thinking & 44.31 & 89.71 & 82.66 & 42.39 & 88.73 & 83.31 & 49.54 & 88.23 & 83.26 & 36.35 & 83.86 & 81.08 & 43.15 & 87.63 & 82.58 \\
\bottomrule
\end{tabular}%
}
\caption{Results on the En$\Rightarrow$Zh WMT22 General Machine Translation task.}
\end{table*}

\begin{table*}[h]
\centering
\resizebox{\textwidth}{!}{%
\begin{tabular}{l ccc ccc ccc ccc ccc}
\toprule
\multirow{2}{*}{Model} & \multicolumn{3}{c}{Conversation} & \multicolumn{3}{c}{E-commerce} & \multicolumn{3}{c}{News} & \multicolumn{3}{c}{Social} & \multicolumn{3}{c}{Average} \\
\cmidrule(lr){2-4} \cmidrule(lr){5-7} \cmidrule(lr){8-10} \cmidrule(lr){11-13} \cmidrule(lr){14-16}
 & BLEU & COMET & KIWI & BLEU & COMET & KIWI & BLEU & COMET & KIWI & BLEU & COMET & KIWI & BLEU & COMET & KIWI \\
\midrule
GPT-4o & \textbf{40.65} & 89.88 & 83.72 & \textbf{38.03} & 88.80 & 84.02 & \textbf{36.70} & 87.35 & 84.00 & 34.06 & 85.12 & 82.63 & \textbf{37.36} & 87.79 & 83.59 \\
DeepSeek-V3 & 39.58 & 89.33 & 83.50 & 37.42 & 88.69 & 84.14 & 36.48 & 87.94 & 84.76 & \textbf{35.08} & 85.22 & 82.91 & 37.14 & 87.79 & 83.83 \\
Gemini-2.0-Flash & 37.88 & 88.46 & 83.36 & 37.47 & \textbf{88.98} & 84.09 & 35.88 & 87.69 & 84.63 & 34.02 & 85.39 & 82.84 & 36.31 & 87.63 & 83.73 \\
\hdashline
OpenAI-o1 & 38.71 & \textbf{90.19} & \textbf{85.55} & 36.25 & 88.73 & \textbf{85.79} & 33.90 & \textbf{88.17} & \textbf{86.69} & 32.18 & \textbf{85.73} & \textbf{84.54} & 35.26 & \textbf{88.21} & \textbf{85.64} \\
DeepSeek-R1 & 36.17 & 89.11 & 83.60 & 34.70 & 88.73 & 84.27 & 33.96 & 87.92 & 84.87 & 32.39 & 85.22 & 82.85 & 34.31 & 87.75 & 83.90 \\
OpenAI-o3-mini & 35.35 & 88.05 & 83.21 & 36.71 & 88.28 & 84.01 & 34.01 & 87.16 & 84.31 & 32.14 & 84.29 & 82.25 & 34.55 & 86.95 & 83.44 \\
Gemini-2.0-Flash-Thinking & 36.64 & 88.04 & 83.00 & 37.87 & 88.97 & 83.98 & 36.18 & 87.81 & 84.71 & 33.84 & 85.00 & 82.59 & 36.13 & 87.45 & 83.57 \\
\bottomrule
\end{tabular}%
}
\caption{Results on the En$\Rightarrow$De WMT22 General Machine Translation task.}
\end{table*}

\begin{table*}[h]
\centering
\resizebox{\textwidth}{!}{%
\begin{tabular}{l ccc ccc ccc ccc ccc}
\toprule
\multirow{2}{*}{Model} & \multicolumn{3}{c}{Conversation} & \multicolumn{3}{c}{E-commerce} & \multicolumn{3}{c}{News} & \multicolumn{3}{c}{Social} & \multicolumn{3}{c}{Average} \\
\cmidrule(lr){2-4} \cmidrule(lr){5-7} \cmidrule(lr){8-10} \cmidrule(lr){11-13} \cmidrule(lr){14-16}
 & BLEU & COMET & KIWI & BLEU & COMET & KIWI & BLEU & COMET & KIWI & BLEU & COMET & KIWI & BLEU & COMET & KIWI \\
\midrule
GPT-4o & \textbf{37.14} & 86.07 & 81.79 & \textbf{34.87} & \textbf{86.15} & 81.37 & \textbf{31.74} & 85.49 & 82.14 & \textbf{32.35} & 85.07 & 81.27 & \textbf{34.02} & 85.69 & 81.64 \\
DeepSeek-V3 & 36.52 & 86.24 & 81.84 & 34.56 & 86.04 & 81.46 & 30.76 & 85.52 & 82.47 & 31.65 & 84.93 & 81.43 & 33.37 & 85.68 & 81.80 \\
Gemini-2.0-Flash & 36.76 & 86.05 & 81.92 & 34.30 & 86.02 & 81.23 & 31.28 & 85.64 & 82.35 & 31.57 & 84.88 & 81.37 & 33.48 & 85.65 & 81.72 \\
\hdashline
OpenAI-o1 & 34.73 & \textbf{86.49} & \textbf{83.40} & 31.91 & 85.95 & \textbf{82.65} & 30.35 & 85.78 & \textbf{83.93} & 30.36 & \textbf{85.22} & \textbf{82.93} & 31.84 & \textbf{85.86} & \textbf{83.23} \\
DeepSeek-R1 & 35.20 & 86.06 & 81.96 & 32.11 & 85.75 & 81.39 & 29.59 & 85.78 & 82.66 & 30.04 & 84.92 & 81.45 & 31.73 & 85.63 & 81.86 \\
OpenAI-o3-mini & 36.16 & 85.66 & 81.54 & 34.20 & 85.59 & 81.26 & 31.07 & 84.71 & 81.30 & 32.02 & 84.66 & 81.00 & 33.36 & 85.16 & 81.28 \\
Gemini-2.0-Flash-Thinking & 36.35 & 86.05 & 81.79 & 33.05 & 85.77 & 81.14 & 31.59 & \textbf{85.83} & 82.42 & 31.48 & 84.76 & 81.28 & 33.12 & 85.60 & 81.66 \\
\bottomrule
\end{tabular}%
}
\caption{Results on the De$\Rightarrow$En WMT22 General Machine Translation task.}
\end{table*}

\begin{table*}[h]
\centering
\resizebox{\textwidth}{!}{%
\begin{tabular}{l ccc ccc ccc ccc ccc}
\toprule
\multirow{2}{*}{Model} & \multicolumn{3}{c}{WMTBio18 En$\Rightarrow$Zh} & \multicolumn{3}{c}{WMTBio18 Zh$\Rightarrow$En} & \multicolumn{3}{c}{WMTBio18 En$\Rightarrow$De} & \multicolumn{3}{c}{WMTBio18 De$\Rightarrow$En} & \multicolumn{3}{c}{Average} \\
\cmidrule(lr){2-4} \cmidrule(lr){5-7} \cmidrule(lr){8-10} \cmidrule(lr){11-13} \cmidrule(lr){14-16}
 & BLEU & COMET & KIWI & BLEU & COMET & KIWI & BLEU & COMET & KIWI & BLEU & COMET & KIWI & BLEU & COMET & KIWI \\
\midrule
GPT-4o & \textbf{40.86} & 86.66 & 84.61 & 25.04 & 84.33 & 84.02 & 24.44 & 86.68 & 84.97 & 29.31 & 85.06 & 82.61 & 29.91 & 85.68 & 84.05 \\
DeepSeek-V3 & 40.58 & 86.67 & 84.66 & 25.73 & 84.42 & 83.95 & 24.75 & 86.85 & 84.96 & 29.37 & 85.01 & 82.78 & 30.11 & 85.74 & 84.09 \\
Gemini-2.0-Flash & 40.83 & \textbf{86.87} & 84.62 & \textbf{26.32} & \textbf{84.50} & 83.74 & 26.47 & 86.72 & 84.66 & \textbf{30.69} & \textbf{85.10} & 82.22 & 31.08 & \textbf{85.80} & 83.81 \\
\hdashline
OpenAI-o1 & 38.84 & 86.63 & \textbf{85.52} & 24.85 & 84.30 & \textbf{84.88} & \textbf{32.29} & \textbf{86.96} & \textbf{86.68} & 28.42 & 85.08 & \textbf{84.05} & \textbf{31.35} & 85.74 & \textbf{85.38} \\
DeepSeek-R1 & 38.43 & 86.68 & 84.45 & 24.34 & 84.22 & 83.70 & 25.14 & 86.94 & 85.01 & 29.34 & 85.04 & 82.79 & 29.31 & 85.72 & 83.99 \\
OpenAI-o3-mini & 39.39 & 86.15 & 84.36 & 25.09 & 83.99 & 83.77 & 22.96 & 86.39 & 84.73 & 29.06 & 84.70 & 82.77 & 29.12 & 85.31 & 83.91 \\
Gemini-2.0-Flash-Thinking & 39.90 & 86.36 & 84.35 & 25.54 & 84.29 & 83.90 & 25.78 & 86.75 & 84.72 & 30.57 & 84.98 & 82.86 & 30.45 & 85.59 & 83.95 \\
\bottomrule
\toprule
\multirow{2}{*}{Model} & \multicolumn{3}{c}{WMTBio19 En$\Rightarrow$Zh} & \multicolumn{3}{c}{WMTBio19 Zh$\Rightarrow$En} & \multicolumn{3}{c}{WMTBio19 En$\Rightarrow$De} & \multicolumn{3}{c}{WMTBio19 De$\Rightarrow$En} & \multicolumn{3}{c}{Average} \\
\cmidrule(lr){2-4} \cmidrule(lr){5-7} \cmidrule(lr){8-10} \cmidrule(lr){11-13} \cmidrule(lr){14-16}
 & BLEU & COMET & KIWI & BLEU & COMET & KIWI & BLEU & COMET & KIWI & BLEU & COMET & KIWI & BLEU & COMET & KIWI \\
\midrule
GPT-4o & 37.58 & 86.75 & 84.53 & 31.41 & 84.42 & 82.72 & 30.49 & 86.29 & 83.55 & 37.25 & 86.91 & 82.35 & 34.18 & 86.09 & 83.29 \\
DeepSeek-V3 & 38.54 & 86.93 & 84.59 & 30.75 & \textbf{84.68} & 83.10 & 34.14 & 86.47 & 83.91 & 37.63 & \textbf{87.66} & 82.29 & 35.26 & 86.44 & 83.48 \\
Gemini-2.0-Flash & \textbf{38.89} & 86.91 & 84.48 & \textbf{32.59} & 84.50 & 82.86 & \textbf{34.59} & \textbf{87.96} & 83.96 & \textbf{37.87} & 86.71 & 82.24 & \textbf{35.98} & \textbf{86.52} & 83.39 \\
\hdashline
OpenAI-o1 & 37.01 & 86.97 & \textbf{85.42} & 28.15 & 84.53 & \textbf{83.50} & 32.29 & 83.52 & \textbf{85.88} & 36.62 & 84.93 & \textbf{83.82} & 33.52 & 85.00 & \textbf{84.66} \\
DeepSeek-R1 & 37.23 & \textbf{87.05} & 84.27 & 28.58 & 84.31 & 82.60 & 32.56 & 83.80 & 83.98 & 36.19 & 84.84 & 82.28 & 33.64 & 85.00 & 83.28 \\
OpenAI-o3-mini & 35.62 & 86.07 & 84.22 & 30.59 & 84.04 & 82.91 & 31.55 & 86.16 & 83.86 & 36.92 & 86.48 & 82.19 & 33.67 & 85.69 & 83.29 \\
Gemini-2.0-Flash-Thinking & 38.48 & 86.43 & 84.20 & 31.78 & 84.48 & 82.88 & 33.55 & 86.82 & 83.89 & 36.74 & 84.57 & 82.25 & 35.14 & 85.57 & 83.31 \\
\bottomrule
\end{tabular}%
}
\caption{Results on the Zh$\Leftrightarrow$En and De$\Leftrightarrow$En Biomedical Machine Translation Task from WMT18 and WMT19 (WMTBio18 and WMTBio19). }
\label{tab:result_wmtbio_updated_averages}
\end{table*}

\begin{table*}[h]
\centering
\resizebox{\textwidth}{!}{%
\begin{tabular}{l ccc ccc ccc ccc ccc ccc}
\toprule
\multirow{2}{*}{Model} 
& \multicolumn{3}{c}{Medical} & \multicolumn{3}{c}{Law} & \multicolumn{3}{c}{IT} 
& \multicolumn{3}{c}{Koran} & \multicolumn{3}{c}{Subtitles} & \multicolumn{3}{c}{Average} \\
\cmidrule(lr){2-4} \cmidrule(lr){5-7} \cmidrule(lr){8-10}
\cmidrule(lr){11-13} \cmidrule(lr){14-16} \cmidrule(lr){17-19}
& BLEU & COMET & KIWI & BLEU & COMET & KIWI & BLEU & COMET & KIWI
& BLEU & COMET & KIWI & BLEU & COMET & KIWI & BLEU & COMET & KIWI \\
\midrule
GPT-4o & 41.89 & 84.23 & 81.89 & 39.00 & 85.06 & 82.38 & 37.23 & 83.52 & 78.37 & 17.65 & \textbf{75.04} & 80.64 & 29.61 & 80.62 & 82.25 & 33.08 & 81.69 & 81.11 \\
DeepSeek-V3 & 41.43 & 84.15 & 81.90 & 38.82 & 84.94 & 82.29 & \textbf{38.16} & \textbf{83.90} & 78.86 & 17.94 & 74.91 & 80.33 & 29.70 & 80.68 & 82.26 & 33.21 & 81.72 & 81.13 \\
Gemini-2.0-Flash & \textbf{44.39} & \textbf{84.49} & 81.76 & \textbf{46.15} & \textbf{85.57} & 82.30 & 37.93 & 83.20 & 78.49 & \textbf{19.70} & 74.97 & 79.71 & \textbf{30.39} & 80.74 & 82.19 & \textbf{35.71} & \textbf{81.79} & 80.89 \\
\hdashline
OpenAI-o1 & 40.44 & 83.88 & \textbf{83.29} & 40.74 & 85.03 & \textbf{83.53} & 36.40 & 83.20 & \textbf{79.75} & 16.76 & 75.03 & \textbf{82.02} & 28.91 & \textbf{80.79} & \textbf{83.65} & 32.65 & 81.59 & \textbf{82.45} \\
DeepSeek-R1 & 40.23 & 83.82 & 81.81 & 39.16 & 84.97 & 82.35 & 36.66 & 83.49 & 78.68 & 17.05 & 74.76 & 80.63 & 28.82 & 80.65 & 82.22 & 32.38 & 81.54 & 81.14 \\
OpenAI-o3-mini & 39.82 & 83.60 & 81.79 & 34.29 & 84.18 & 81.96 & 37.13 & 82.63 & 78.20 & 15.52 & 73.54 & 80.19 & 28.39 & 80.06 & 82.06 & 31.03 & 80.80 & 80.84 \\
Gemini-2.0-Flash-Thinking & 43.36 & 84.14 & 81.70 & 44.51 & 85.10 & 82.13 & 37.26 & 82.94 & 78.40 & 19.57 & 75.03 & 79.79 & 30.38 & 80.78 & 82.20 & 35.02 & 81.60 & 80.84 \\
\bottomrule
\end{tabular}
}
\caption{Results on the Dn$\Rightarrow$En Multi-Domain dataset.}
\label{tab:result_multi_domain}
\end{table*}

\begin{table*}[h]
\centering
\resizebox{\textwidth}{!}{%
\begin{tabular}{l ccc ccc ccc ccc ccc}
\toprule
\multirow{2}{*}{Model} & \multicolumn{3}{c}{Test-1} & \multicolumn{3}{c}{Test-2} & \multicolumn{3}{c}{Valid-1} & \multicolumn{3}{c}{Valid-2} & \multicolumn{3}{c}{Average} \\
\cmidrule(lr){2-4} \cmidrule(lr){5-7} \cmidrule(lr){8-10} \cmidrule(lr){11-13} \cmidrule(lr){14-16}
& BLEU & COMET & KIWI & BLEU & COMET & KIWI & BLEU & COMET & KIWI & BLEU & COMET & KIWI & BLEU & COMET & KIWI \\
\midrule
GPT-4o & 20.49 & 78.44 & 76.50 & 16.12 & 77.55 & 77.82 & 21.08 & 78.91 & 77.31 & 12.49 & 75.38 & 76.91 & 17.54 & \textbf{77.57} & 77.13 \\
DeepSeek-V3 & 15.66 & 77.47 & 77.84 & \textbf{16.25} & 77.14 & 77.49 & 19.57 & \textbf{78.94} & \textbf{77.48} & 11.87 & 75.28 & 77.05 & 16.81 & 77.48 & 77.17 \\
Gemini-2.0-Flash & \textbf{22.11} & 78.25 & 75.94 & \textbf{16.25} & 77.14 & 77.49 & \textbf{21.42}& 78.70 & 76.70 & \textbf{12.70} & 75.11 & 76.34 & \textbf{18.12} & 77.30 & 76.62 \\
\hdashline
OpenAI-o1 & 19.05 & \textbf{78.85} & \textbf{78.00} & 16.04 & \textbf{77.76} & \textbf{78.66} & 18.03 & 78.13 & 76.93 & 12.65 & \textbf{75.52} & \textbf{77.44} & 16.44 & 77.56 & \textbf{77.76} \\
DeepSeek-R1 & 15.34 & 77.29 & 75.72 & 12.68 & 77.04 & 77.23 & 14.35 & 77.67 & 76.59 & 10.09 & 74.55 & 76.33 & 13.12 & 76.64 & 76.47 \\
OpenAI-o3-mini & 20.63 & 77.77 & 76.02 & 15.57 & 76.69 & 77.38 & 21.11 & 78.35 & 76.97 & 11.74 & 74.69 & 76.17 & 17.26 & 76.88 & 76.63 \\
Gemini-2.0-Flash-Thinking & 21.69 & 78.78 & 76.71 & 15.75 & 76.98 & 77.43 & 21.14 & 78.17 & 75.90 & 12.48 & 74.75 & 76.27 & 17.77 & 77.17 & 76.58 \\
\bottomrule
\end{tabular}%
}
\caption{Results on the Zh$\Rightarrow$En GuoFeng Webnovel dataset (Literary domain).}
\label{tab:result_webnovel}
\end{table*}

\begin{table*}[h]
\centering
\resizebox{\textwidth}{!}{%
\begin{tabular}{l ccc ccc ccc ccc ccc}
\toprule
\multirow{2}{*}{Model} & \multicolumn{3}{c}{WMT22 En$\Rightarrow$Zh} & \multicolumn{3}{c}{WMT22 Zh$\Rightarrow$En} & \multicolumn{3}{c}{WMT22 En$\Rightarrow$De} & \multicolumn{3}{c}{WMT22 De$\Rightarrow$En} & \multicolumn{3}{c}{Average} \\
\cmidrule(lr){2-4} \cmidrule(lr){5-7} \cmidrule(lr){8-10} \cmidrule(lr){11-13} \cmidrule(lr){14-16}
& BLEU & COMET & KIWI & BLEU & COMET & KIWI & BLEU & COMET & KIWI & BLEU & COMET & KIWI & BLEU & COMET & KIWI \\
\midrule
DeepSeek-V3 & 44.52 & \textbf{88.21} & 83.19 & 25.21 & 83.11 & 79.95 & 36.89 & 87.78 & 83.83 & 33.06 & 85.66 & 81.80 & 34.92 & 86.19 & 82.19 \\
GPT-4o & \textbf{45.64} & 87.77 & 82.74 & 25.11 & \textbf{83.22} & 79.95 & \textbf{37.07} & 87.77 & 83.60 & \textbf{33.71} & 85.68 & 81.64 & \textbf{35.38} & 86.11 & 81.98 \\
Gemini-2.0-Flash & 45.50 & 88.04 & 82.84 & \textbf{26.12} & 82.94 & 79.56 & 36.16 & 87.63 & 83.74 & 33.20 & 85.63 & 81.71 & 35.25 & 86.06 & 81.96 \\
\hdashline
OpenAI-o1 & 40.84 & 88.10 & \textbf{84.10} & 22.42 & 83.08 & \textbf{80.87} & 34.87 & \textbf{88.18} & \textbf{85.64} & 31.53 & \textbf{85.84} & \textbf{83.22} & 32.42 & \textbf{86.30} & \textbf{83.46} \\
DeepSeek-R1 & 36.88 & 87.33 & 82.96 & 20.68 & 82.48 & 79.57 & 34.14 & 87.73 & 83.91 & 31.37 & 85.61 & 81.86 & 30.77 & 85.79 & 82.07 \\
OpenAI-o3-mini & 42.67 & 87.25 & 82.52 & 24.09 & 82.82 & 79.89 & 34.46 & 86.94 & 83.45 & 33.08 & 85.14 & 81.27 & 33.58 & 85.54 & 81.78 \\
Gemini-2.0-Flash-Thinking & 44.04 & 87.61 & 82.58 & 24.83 & 82.71 & 79.51 & 36.12 & 87.46 & 83.58 & 32.75 & 85.59 & 81.66 & 34.44 & 85.84 & 81.83 \\
\bottomrule
\end{tabular}%
}
\caption{Results on the Zh$\Leftrightarrow$En and De$\Leftrightarrow$En WMT22 General Machine Translation task.}
\label{tab:result_wmt22}
\end{table*}

\label{appendix:results}

\section{Prompt}
The prompt for calculating MQM scores is provided in Figure \ref{fig:mqm-prompt}.
The prompt for evaluating translation difficulty is shown in the Figure \ref{fig:difficulty_eval}.

\begin{figure*}[t]
\centering
\begin{tcolorbox}[colback=gray!10, colframe=black, width=\textwidth]
\small
\setstretch{1.2}

\noindent{You are a professional translation quality evaluator following the MQM (Multidimensional Quality Metrics) framework.}

\textbf{Task Instructions:}
\begin{enumerate}
    \item Compare the Prediction against both the Source and the Reference.
    \item Identify up to five of the most serious Errors for each translation sentence, using the MQM error types listed below.
    \item Assign exactly one severity level to each error.
    \item Special handling for two specific error types:
    \begin{itemize}
        \item \textbf{Source Error}: Errors present in the Source text itself.
        \item \textbf{Non-translation Error}: Translation is unassessable and unrelated to the Source.
    \end{itemize}
    \item If no errors are found, return an empty JSON list [].
    \item Only output a valid JSON object. Do not include any additional text, comments, or explanations.
\end{enumerate}

\textbf{MQM Error Types (See Table~\ref{tab:mqm-hierarchy} for the complete hierarchy):}
\begin{itemize}
    \item Mistranslation: Incorrect translation that alters or distorts meaning.
    \item Addition: Insertion of information or emotion not present in the source.
    \item Under-translation: Partial omission of relevant or necessary information.
    \item Omission: Complete exclusion of content present in the source.
    \item Untranslated: Source text is copied without being translated.
\end{itemize}

\textbf{Severity Levels:} 
\begin{itemize}
    \item Minor: Slight impact on readability or style; meaning remains clear.
    \item Major: Significant impact on usability, comprehension, or meaning.
\end{itemize}
\textbf{Output Format:}
\begin{verbatim}
{
  "errors": [
    {
      "error_type": "Mistranslation",
      "severity": "Major",
      "explanation": "The correct translation should be ... "
    }
  ]
}
\end{verbatim}

\textbf{Source:} \texttt{\{source\_text\}} \\
\textbf{Reference:} \texttt{\{reference\_text\}} \\
\textbf{Prediction:} \texttt{\{prediction\_text\}}

\end{tcolorbox}
\caption{Full prompt used to calculate MQM scores with DeepSeek-V3. }
\label{fig:mqm-prompt}
\end{figure*}

\begin{figure*}[t]
\centering
\begin{tcolorbox}[colback=gray!10, colframe=black, width=\textwidth]

Your task is to assess the difficulty of translating a given \texttt{\{src\_lang\}} sentence into \texttt{\{tgt\_lang\}}. Please evaluate the difficulty based on the following criteria and output the result in JSON format, with the key "level":
\begin{enumerate}
\item  Sentence complexity: Determine if the sentence is a simple sentence, a compound sentence, or includes subordinate clauses and other complex structures.
\item  Vocabulary difficulty: Assess whether the sentence contains commonly used words or specialized terms or slang.
\item  Grammar differences: Analyze if the sentence's grammatical structure is similar to or differs significantly from \texttt{\{tgt\_lang\}}.
\item  Contextual understanding: Consider whether understanding specific cultural contexts or background knowledge is necessary for accurate translation.
\end{enumerate}

Provide a difficulty level (1-5), with 1 being the easiest and 5 being the most difficult.
And output the difficulty level in the following JSON format:
\begin{verbatim}
{
    "level": "difficulty level"
}
\end{verbatim}

Here is the \texttt{\{src\_lang\}} sentence: \texttt{\{src\_text\}}

\end{tcolorbox}
\caption{Full prompt used for evaluating translation difficulty with DeepSeek-V3.}
\label{fig:difficulty_eval}
\end{figure*}

\end{document}